\documentclass[runningheads]{llncs}

\usepackage{cleveref}
\usepackage[T1]{fontenc}
\usepackage{graphicx}

\usepackage{amssymb,amsfonts}
\usepackage{bm}
\usepackage{array}
\usepackage{tabularx}
\usepackage{makecell}
\usepackage{xcolor}
\definecolor{CrashRed}{RGB}{192,0,0}
\definecolor{CrashGreen}{RGB}{0,176,80}

\begin{document}
\title{CrashChat: A Multimodal Large Language Model for Multitask Traffic Crash Video Analysis}

\titlerunning{CrashChat: MLLM for Crash Video Analysis}

\author{Kaidi Liang\inst{1}\orcidID{0009-0001-9129-2744} \and
Ke Li\inst{1}\orcidID{0009-0001-4958-3302} \and
Xianbiao Hu\inst{2}\orcidID{0000-0002-0149-1847} \and
Ruwen Qin\inst{1}\orcidID{0000-0003-2656-8705}}

\authorrunning{Liang et al.}

\institute{
Stony Brook University, Department of Civil Engineering, Stony Brook, NY 11794, USA\\
\email{\{kaidi.liang,ke.li,ruwen.qin\}@stonybrook.edu}
\and
The Pennsylvania State University, Department of Civil and Environmental Engineering, University Park, PA 16802-1408, USA\\
\email{xbhu@psu.edu}
}
\maketitle              
\begin{abstract}
Automating crash video analysis is essential to leverage the growing availability of driving video data for traffic safety research and accountability attribution in autonomous driving. Crash video analysis is a challenging multitask problem due to the complex spatiotemporal dynamics of crash events in video data and the diverse analytical requirements involved. It requires capabilities spanning crash recognition, temporal grounding, and high-level video understanding. Existing models, however, cannot perform all these tasks within a unified framework, and effective training strategies for such models remain underexplored. To fill these gaps, this paper proposes CrashChat, a multimodal large language model (MLLM) for multitask traffic crash analysis, built upon VideoLLaMA3. CrashChat acquires domain-specific knowledge through instruction fine-tuning and employs a novel multitask learning strategy based on task decoupling and grouping, which maximizes the benefit of joint learning within and across task groups while mitigating negative transfer. Numerical experiments on consolidated public datasets demonstrate that CrashChat consistently outperforms existing MLLMs across model scales and traditional vision-based methods, achieving state-of-the-art performance. It reaches near-perfect accuracy in crash recognition, a 176\% improvement in crash localization, and a 40\% improvement in the more challenging pre-crash localization. Compared to general MLLMs, it substantially enhances textual accuracy and content coverage in crash description and reasoning tasks, with 0.18-0.41 increases in BLEU scores and 0.18-0.42 increases in ROUGE scores. Beyond its strong performance, CrashChat is a convenient, end-to-end analytical tool ready for practical implementation. The dataset and implementation code for CrashChat are available at https://github.com/Liangkd/CrashChat.

\end{abstract}
\section{Introduction}

Video data are increasingly recognized as a readily accessible and informative source for traffic crash analysis, especially for autonomous vehicles. For instance, retrieving and analyzing crash footage captured by in-vehicle cameras can provide objective evidence about adverse events, facilitating timely and accurate crash reporting. Given the vast volume of captured video data and the complexity of crash events, automating crash-related video analysis has become a critical need. Specifically, a tool is needed to automatically identify anomalous segments from massive amounts of video data, ground them in their temporal evolution, and generate a semantic understanding of the events.

\begin{figure}
\centering
\includegraphics[width=\textwidth]{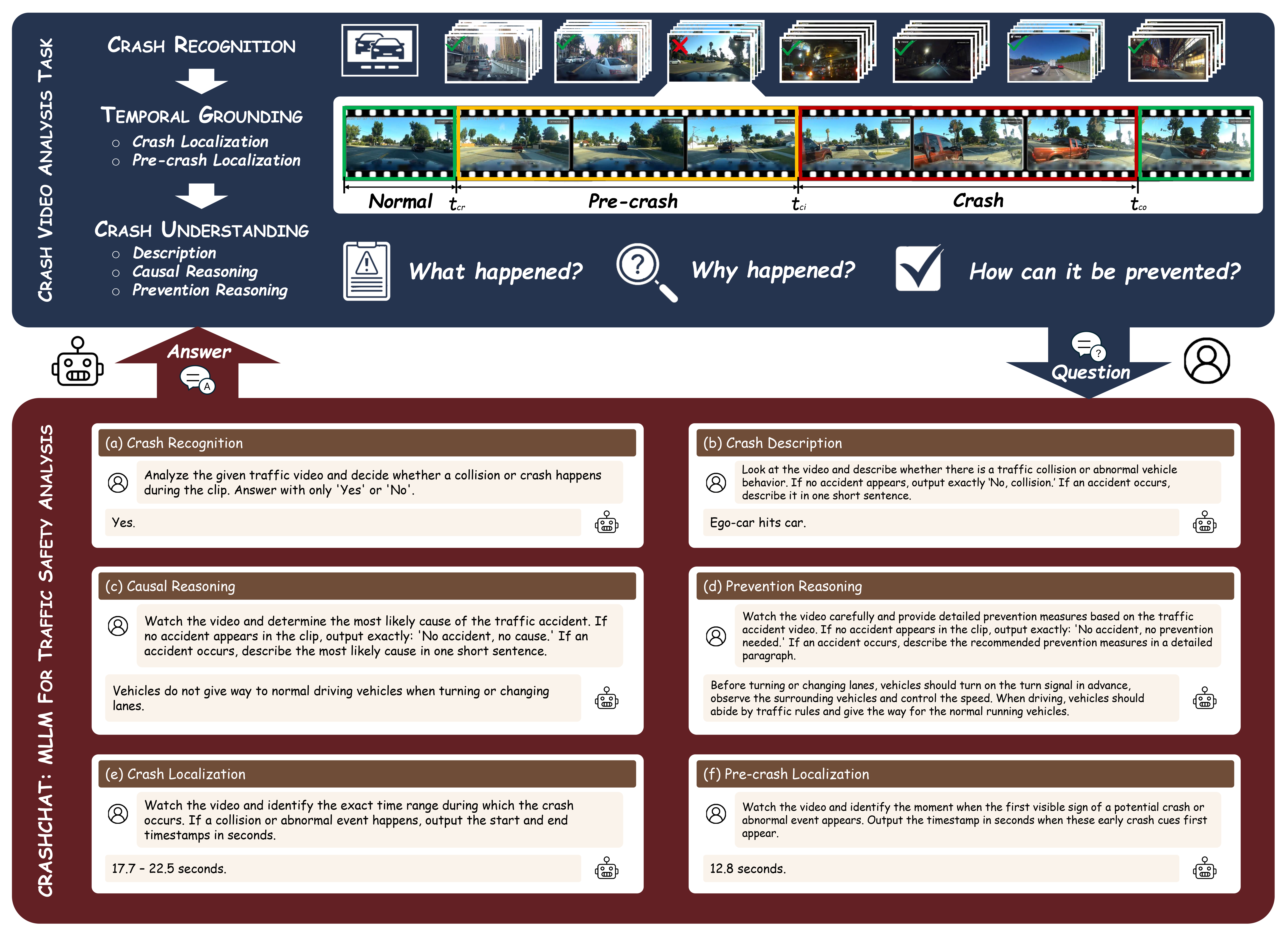}
\caption{CrashChat - a multitask multimodal large language model performing six core tasks in support of crash video analysis in a unified way} \label{figure1}
\end{figure}

Crash video analysis is inherently a multitask problem. The desired capabilities, including crash recognition, temporal grounding, and understanding, can be enabled by six core tasks, as illustrated in Fig. \ref{figure1}. Crash recognition determines whether the input video contains a crash event. Crash localization identifies the precise interval during which the crash occurred, while pre-crash localization further determines when visual cues first appeared, signaling that a crash is imminent. Crash description provides a structured narrative of the event. Causal reasoning infers the underlying causes of the crash, and prevention reasoning identifies conditions or actions that could have avoided it.

While focusing on different aspects of crash video analysis, the six tasks are intertwined. Crash recognition narrows down to the target segment, thereby enhancing the efficiency of temporal grounding. Further unfolding the identified segment into normal, pre-crash, and crash phases assists in distinguishing causal actions from subsequent effects. Conversely, the improved understanding of crash events will further enhance crash recognition. Therefore, a multitask framework can perform better than its monotask counterparts. Positive outcomes from multitasking rely on a good understanding of the relationship among those tasks and appropriate designs of the training strategy \cite{ding_2023_etrnlp}. Yet, this remains largely unexplored for crash video analysis, where task interactions are complex.


Multimodal large language models (MLLMs) show strong potential for building the desired capabilities for crash video analysis. Recent studies demonstrate their impressive ability to describe and reason about crash events, drawing increasing attention from the research community \cite{safeplug_2025,accidentgpt_2023,echo_2025,106k_2025}. Although crash recognition and temporal grounding tasks are traditionally formulated as computer vision problems (e.g., \cite{karim2023amnet,orlova2025simplifying,yao2019unsupervised}), MLLMs have also shown greater promise in these tasks \cite{safeplug_2025,echo_2025,106k_2025}. This advantage stems from their ability to unfold the details of crash events over time and in description. However, there is currently no unified MLLM model capable of addressing the full spectrum of tasks required for comprehensive crash video analysis.

As one of the most advanced MLLMs, VideoLLaMA3 \cite{videollama3_2025} achieves state-of-the-art (SOTA) performance. Its vision-centric training paradigm, vision encoder adaptation, vision-language alignment, multi-task fine-tuning, and video-centric fine-tuning, along with differential frame pruning, a unique design for video handling, creates an exciting opportunity for crash video analysis. Despite these methodological advantages, VideoLLaMA3 has not yet been evaluated by this critical application domain.

To fill existing gaps and advance the frontier of MLLMs for crash video analysis, this paper makes the following contributions: (\textbf{i}) a multitask learning approach designed to effectively inject the comprehensive knowledge of crash video analysis into VideoLLaMA3; ($\textbf{ii}$) an MLLM capable of unified crash recognition, temporal grounding, and understanding across diverse scenarios; and ($\textbf{iii}$) a comprehensive evaluation that provides the first benchmarking of MLLMs for end-to-end crash video analysis.

The remainder of the paper is organized as follows. \Cref{sec:Literature} reviews related work, \Cref{sec:Methodology} presents the proposed approach, \Cref{sec:Implement} describes the experimental setup, \Cref{sec:results} reports the evaluation results, and \Cref{sec:Concusion} concludes the paper with a summary of learned insights and future research directions.

\section{Literature Review}
\label{sec:Literature}

Current studies primarily contribute to crash video analysis from three aspects: datasets, vision-based approaches, and MLLM-based methods.

\subsection{Traffic Crash Datasets}
\label{subsec:Literature_Datasets}

Crash video datasets are generally divided into two categories according to their target applications: (1) Traffic Crash Detection (TCD) (e.g., \cite{fang2024abductive,fang2019dada,lv2021wsal,yao2019unsupervised,106k_2025}) and (2) Traffic Crash Anticipation (TCA) (e.g., \cite{karim2023amnet,moura2025nexar}). TAD datasets focus more on crash description and reasoning, often providing the cause of the crash (e.g., \cite{fang2024abductive,yao2022dota,106k_2025}) and, in some cases, possible preventive actions \cite{fang2024abductive}. In contrast, TCA datasets comprise both positive and negative video clips and emphasize crash recognition and temporal localizations of the crash and pre-crash phases \cite{moura2025nexar}. 


Recently, several datasets have begun incorporating more diverse multimodal information to better align with MLLM frameworks, such as MM-AU \cite{fang2024abductive}, SafePLUG \cite{safeplug_2025} and AV-TAU \cite{echo_2025}. However, from the perspective of traffic crash analysis, there remains a lack of a unified dataset that can support all tasks required for comprehensive crash video analysis.

\subsection{Vision-based Approaches}

Traditionally, crash video analysis is formulated as computer vision problems (e.g., \cite{karim2023amnet,orlova2025simplifying,yao2019unsupervised}). This approach relies on extracting features from video data to learn the spatiotemporal dynamics of traffic participants in driving scenes to support downstream tasks. For example, Yao et al. \cite{yao2019unsupervised} predicted future ego-motion trajectories from dashcam videos to detect abnormal events. Karim et al. \cite{karim2023amnet} introduced an attention-guided multi-stream feature fusion network to identify dangerous traffic agents, achieving improved performance. Overall, these studies use video data as the sole input and primarily address anomaly detection, temporal localization, and spatial grounding.

Recently, some studies have begun integrating text information with video data to enhance task performance \cite{fang2022cognitive,liang2024textdriven,sun2025earccpmnet}. Although these methods represent a shift towards multimodal inputs, they have not yet fully explored the potential of vision-language models in crash description or reasoning.

\subsection{MLLM-based Approaches}

The advancement of MLLMs opens up new opportunities, introducing cross-modal reasoning into crash video analysis. Some studies directly integrate a pre-trained Large Language Model (LLM) with vision-based models \cite{jaradat2025near_miss,shi2025scvlm,yao2025ifinder}. SCVLM \cite{shi2025scvlm} proposes a typical paradigm where the model first classifies collision videos and then uses an LLM to generate collision descriptions. Since the LLM is not trained on traffic crash domain data, its domain knowledge and crash reasoning ability remain limited.

On the other hand, several studies inject domain knowledge into MLLMs by fully training or fine-tuning them on crash video datasets \cite{safeplug_2025,echo_2025,106k_2025}. In contrast, such trained MLLMs exhibit stronger capabilities in crash description and reasoning, and often integrate richer multimodal information \cite{echo_2025} or demonstrate deeper fine-grained reasoning abilities \cite{safeplug_2025}. However, there is still no unified framework that can simultaneously address all core tasks, particularly the highly challenging pre-crash localization problem.

\section{Methodology}
\label{sec:Methodology}

\begin{figure}[ht]
\includegraphics[width=\textwidth]{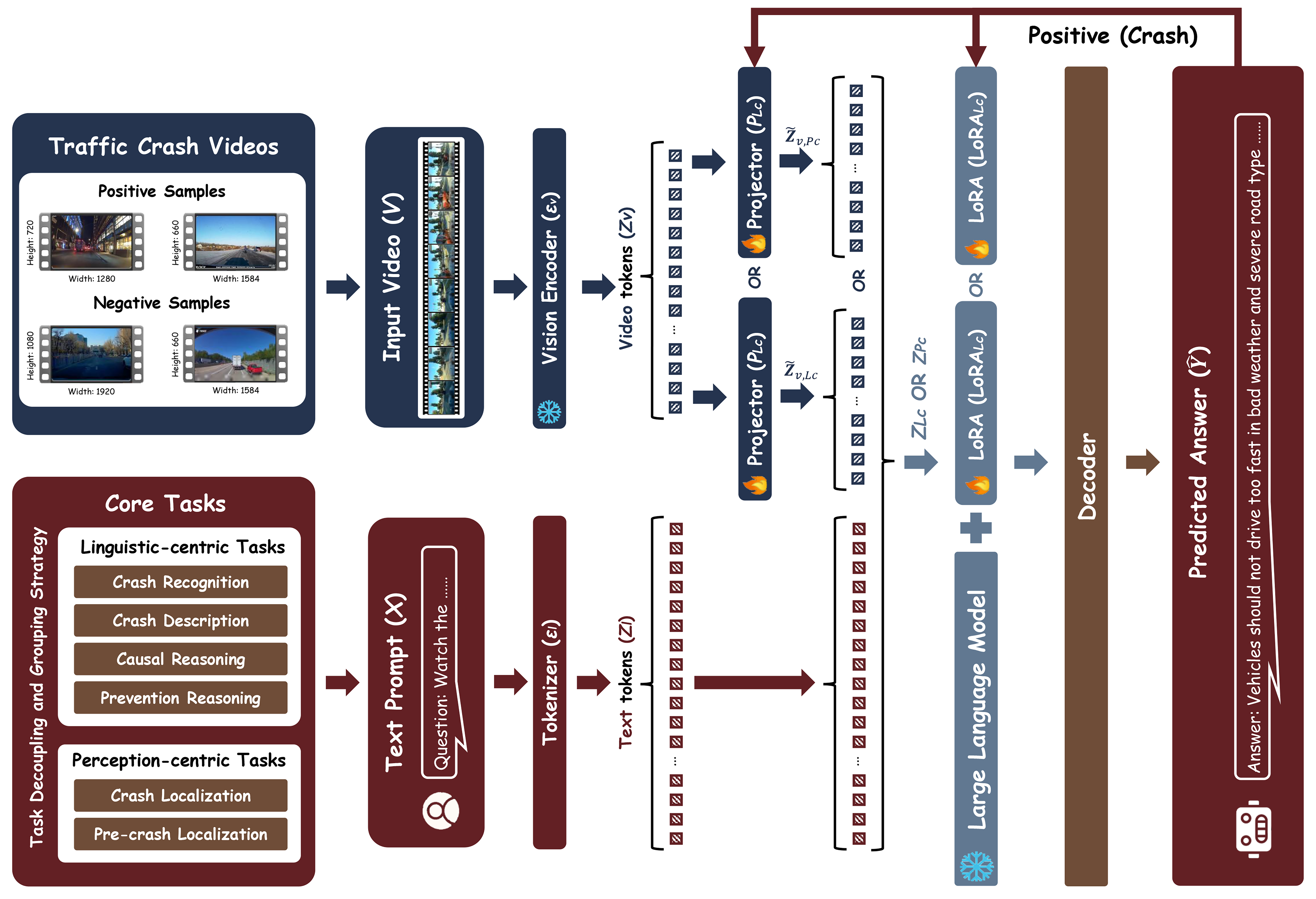}
\caption{Overview of the CrashChat's model architecture. Video and text inputs are encoded into tokens and jointly processed by the LLM to generate task outputs, with temporal grounding applied only to detected positive samples.} 
\label{fig2}
\end{figure}

CrashChat is built upon the VideoLLaMA3 framework and fine-tuned on traffic crash video datasets using LoRA-based instruction tuning. We propose a task decoupling and grouping strategy for optimizing the CrashChat's multitasking performance, as illustrated in Fig.~\ref{fig2}.

\subsection{Task Decoupling and Grouping}
\label{subsec:Task Decoupling and Grouping Strategy}

Prior work has shown that tasks tend to benefit from joint training when their optimization gradients are aligned, while conflicting gradients often lead to negative transfer \cite{standley2020which}. Some tasks for crash video analysis are perception-oriented, requiring the MLLM to learn frame-level semantics and detect motion changes across frames, while others are linguistic-centric, emphasizing abstract semantic understanding and high-level reasoning. This fact has motivated decoupling the six tasks into two groups: linguistic-centric tasks including (a) crash recognition, (b) crash description, (c) causal reasoning, and (d) prevention reasoning; and perception-centric tasks comprising (e) crash localization and (f) pre-crash localization.

\subsection{Model Architecture}

CrashChat is designed to handle these two groups of tasks differently. As shown in Fig. \ref{fig2}, it integrates an efficient vision encoder ($\mathcal{E}_\mathrm{v}$), a text tokenizer ($\mathcal{E}_\mathrm{l}$), two cross-modal projection modules ($\mathcal{P}_\mathrm{Lc}$ and $\mathcal{P}_\mathrm{Pc}$), and an LLM with two Low-Rank Adapters ($\mathrm{LoRA}_\mathrm{Lc}$ and $\mathrm{LoRA}_\mathrm{Pc}$). $\mathcal{P}_\mathrm{Lc}$ and $\mathrm{LoRA}_\mathrm{Lc}$ focus on linguistic-centric tasks, whereas $\mathcal{P}_\mathrm{Pc}$ and $\mathrm{LoRA}_\mathrm{Pc}$ are dedicated to perception-centric tasks.

CrashChat takes a pair of inputs, including a video ($\mathbf{V}$), and a question in form of text prompt ($\mathbf{X}$), to predict the answer to that question ($\widehat{\mathbf{Y}}$). Specifically, the vision encoder converts the input video into a sequence of video token embeddings:
\begin{equation}
\mathbf{Z}_\mathrm{v}=\mathcal{E}_\mathrm{v}(\mathbf{V})=\{\mathbf{z}_{\mathrm{v},i}\},\quad \mathbf{z}_{\mathrm{v},i}\in\mathbb{R}^{d_v}.
\end{equation}

The text tokenizer converts the text prompt into a sequence of text tokens, and the LLM's embedding layer further transforms them into embeddings:
\begin{equation}
\mathbf{Z}_\mathrm{l}=\mathcal{E}_\mathrm{l}(\mathbf{X})=\{\mathbf{z}_{\mathrm{l},j}\},\quad \mathbf{z}_{\mathrm{l},j}\in\mathbb{R}^{d_l}.
\end{equation}

The two learnable projectors respectively map video token embeddings into the space of language embeddings:
\begin{equation}
\widetilde{\mathbf{Z}}_\mathrm{v,m}
= \mathcal{P}_\mathrm{m}(\mathbf{Z}_\mathrm{v})
= \{\widetilde{\mathbf{z}}_{\mathrm{v,m},i}\},
\quad \widetilde{\mathbf{z}}_{\mathrm{v,m},i} \in \mathbb{R}^{d_l},
\; \mathrm{m} \in \{\mathrm{Lc}, \mathrm{Pc}\},
\end{equation}
where $\widetilde{\mathbf{Z}}_\mathrm{v,Lc}$ and $\widetilde{\mathbf{Z}}_\mathrm{v,Pc}$ are learned representations for linguistic-centric and perception-centric tasks, respectively. They are individually concatenated with the language embedding to form the multimodal inputs,
\begin{equation}
    \mathbf{Z}_\mathrm{m} =  [\widetilde{\mathbf{Z}}_{\mathrm{v,m}}; \mathbf{Z}_{\mathrm{l}}],\quad \mathrm{m} \in \{\mathrm{Lc}, \mathrm{Pc}\},
\end{equation}
which are then respectively fed into the LLM adapted for the two groups of tasks via LoRA with low-rank updates: 
\begin{equation}
    \mathbf{W}_{\mathrm{adt,m}} = \mathbf{W} + \mathbf{B}_\mathrm{m}\mathbf{A}_\mathrm{m}, \; 
    \mathbf{A}_\mathrm{m} \in \mathbb{R}^{r \times d_l}, \ \mathbf{B}_\mathrm{m} \in \mathbb{R}^{d_l \times r}, \ r \ll d_l,\; \mathrm{m} \in \{\mathrm{Lc}, \mathrm{Pc}\},
\end{equation}
where $\mathbf{W}$ remains frozen, $\mathbf{A}_\mathrm{Lc}$ and $\mathbf{B}_\mathrm{Lc}$ are learnable weights for handling the linguistic-centric tasks, and $\mathbf{A}_\mathrm{Pc}$ and $\mathbf{B}_\mathrm{Pc}$ are for perception-centric tasks. The LoRA-based instruction fine-turing allows the MLLM to efficiently learn traffic crash domain knowledge without altering the general video-language capabilities inherited from VideoLLaMA3.

Fig. \ref{fig2} show that $\mathbf{W}_\mathrm{adt,Lc}$ is applied to the multimodal input $\mathbf{Z}_\mathrm{m}$, independent of the question type (i.e., $\forall\, \mathrm{m}\in\{\mathrm{Lc, Pc}\}$). If the question is a linguistic-centric task (i.e., $\mathbf{X}\in\{a, b, c, d\}$), the fine-tuned LLM directly generates the final answer to that question. In contrast, for a perception-centric task (i.e.,  $\mathbf{X}\in\{e, f\}$), an additional step is introduced if the input video has been predicted as one containing a crash event. That is, $\mathbf{W}_\mathrm{adt,Pc}$ is applied to $\mathbf{Z}_\mathrm{Pc}$ again to refine the answer to the question.

\subsection{Training Multitask Models}
\label{subsec: Training Multitask Models}

Learnable parameters for the two sets of projectors and LoRAs, shown in Fig. \ref{fig2}, are optimized via two multitask learning processes. We explicitly decouple the training of perception-centric tasks from linguistic-centric tasks to mitigate the inherent conflict between their learning objectives. However, we conjecture that jointly training these two types of tasks can benefit linguistic-centric performance. Accordingly, in training the linguistic-centric tasks, perception-centric tasks are incorporated as auxiliary objectives to enhance multimodal representation learning, while avoiding direct optimization interference with temporal grounding.

\section{Implementation Details}
\label{sec:Implement}

We consolidate a multi-source public dataset to train the proposed model and evaluate it using selected metrics.

\subsection{Dataset Construction}

As discussed in Section \ref{subsec:Literature_Datasets}, there is no unified dataset that can support all tasks required for comprehensive crash video analysis. The MM-AU dataset \cite{fang2024abductive} provides the most extensive crash-related annotations for this study, supporting model training and evaluation with information including crash start and end timestamps, the start timestamp of the pre-crash phase, crash descriptions, causes, and prevention measures. However, MM-AU has very limited negative samples and primarily consists of short video clips. To address these limitations, this study supplements the 11,322 positive videos in MM-AU with the 750 negative samples from Nexar \cite{moura2025nexar} and 6,313 normal driving videos sampled from D2City \cite{d2city}, resulting in a consolidated dataset of 18,385 videos. The Nexar dataset consists of longer videos, with an average duration of 40 seconds, collected from real-world driving scenes in the United States. D2City, collected by DiDi's platform, contains diverse samples. 
In total, the dataset comprises 96,184 video-QA pairs, with six QA pairs per positive video and four per negative video.

The dataset then is split into train, validation, and test sets with a ratio of 8:1:1, while ensuring that the proportions of positive and negative samples in each subset remain consistent with the overall dataset.

\subsection{Performance Metrics}

Performance metrics are chosen to evaluate the models across various tasks for crash video analysis.

\textbf{Recognition:} Crash recognition is formulated as a video-level classification problem. Therefore, Recall, Precision, and F1 score at the class level are used to evaluate the model’s ability to classify input videos as abnormal or normal.

\textbf{Temporal Grounding:}
Pre-crash and crash localization divides a positive video into phases along the temporal dimension and is therefore evaluated using segmentation metrics, including the IoU between predicted and ground-truth time intervals, as well as Average Precisions(AP) at multiple thresholds (AP@30, AP@50, and AP@70).

The boundary between the normal and pre-crash phases can be ambiguous and is often challenging for human annotators to identify precisely. Although annotation guidelines are provided and discrepancies are typically resolved through discussion among annotators, a degree of subjectivity is unavoidable. To account for this uncertainty, this study extends the human-annotated pre-crash starting point by adding a 0.5-second interval ($\delta$) preceding it, resulting in an interval-based annotation for performance evaluation. The interval is added before the annotated start time, rather than after, because experimental studies indicate that crash anticipation algorithms can recognize crash risk earlier than human observers \cite{li_ictd_2023_gaze}. Therefore, the IoU for the pre-crash localization is calculated as: 
\begin{equation}
\mathrm{IoU}_{\mathrm{pre\mbox{-}crash}} =
\left\{
\begin{array}{ll}
1, & \mbox{if }\hat{t}_{\mathrm{ar}}\in[t_{\mathrm{ar}}-\delta,\;t_{\mathrm{ar}}];\\
(\hat{t}_{\mathrm{ar}}-t_{\mathrm{ai}})\,/\,(t_{\mathrm{ar}}-t_{\mathrm{ai}}),
& \mbox{if }\hat{t}_{\mathrm{ar}}\in(t_{\mathrm{ar}},\;t_{\mathrm{ai}});\\
0, & \mbox{if }\hat{t}_{\mathrm{ar}}\in[0,\;t_{\mathrm{ar}}-\delta)\cup[t_{\mathrm{ai}},\;T).\\
\end{array}
\right.
\end{equation}
where $t_\mathrm{ar}$ is the starting time of the pre-crash phase, $t_\mathrm{ai}$ and

\textbf{Understanding:} Crash understanding is jointly achieved by crash description, causal reasoning, and preventive reasoning, which are linguistic tasks evaluated using  BLEU, ROUGE, and BERT.

\section{Results and Discussions}
\label{sec:results}

Numerical experiments are conducted to validate the proposed design and training strategy for CrashChat by evaluating CrashChat's performance. In addition, a representative qualitative example is presented to provide further insight into the model's behavior and practical effectiveness.

\subsection{Selection of the Baseline MLLM}

We select VideoLLaMA3 as the baseline model for developing CrashChat. To justify this choice, we compare VideoLLaMA3 against a set of representative MLLMs with diverse architectural designs and parameter scales, as summarized in \Cref{tab:sota_mllm}. Specifically, the comparison includes two additional widely used lightweight MLLMs (TimeChat and LLaMA-ViD (7B)), one moderate-size model (LLaMA-ViD (13B)), and one large-scale model (Qwen3-VL (32B)). The results show that VideoLLaMA3 offers a favorable balance between computational efficiency and performance, achieving consistently strong results across all tasks, even though it is not the top-performing model in every metric. 

\begin{table}[ht]
\centering
\caption{Comparison of MLLMs for choosing the baseline model. $\dagger$ indicates the best performance among light-weight (7B) models. \textbf{Bold} values indicate the best performance, while \underline{underlined} values indicate the second-best performance.}
\begin{tabularx}{\textwidth}{|
    >{\raggedright\arraybackslash}p{2.2cm}|
    *{3}{>{\raggedleft\arraybackslash}X}|
    >{\raggedleft\arraybackslash}X|
    >{\raggedleft\arraybackslash}X|
}
\hline

& VideoLLaMA3
& TimeChat
& LLaMA-ViD
& LLaMA-ViD
& Qwen3-VL\\
\hline
Num. Params
& 7B
& 7B
& 7B
& 13B
& 32B\\
\hline
\multicolumn{6}{|l|}{\textit{Crash Recognition}} \\
\hline
Rec & $^\dagger$\textbf{0.6732} & 0.4978 & 0.5799  & 0.5761 & \underline{0.6701} \\
Pre & $^\dagger$\textbf{0.7332}  & 0.3779  & 0.6651 & 0.6520 & \underline{0.7324}\\
F1  & $^\dagger$\textbf{0.5791}  & 0.3150 & 0.5612  & 0.5578 & \underline{0.5754} \\
\hline
\multicolumn{6}{|l|}{\textit{Crash Description}} \\
\hline
BLEU  & $^\dagger$\underline{0.0221} & 0.0005 & 0.0015 & 0.0004 & \textbf{0.0506}\\
ROUGE & $^\dagger$\underline{0.3805} & 0.2078 & 0.3688 & 0.3656 & \textbf{0.4296}\\
BERT  & 0.9099 & 0.8769 & $^\dagger$\textbf{0.9281} & \underline{0.9280} & 0.9275\\
\hline
\multicolumn{6}{|l|}{\textit{Causal Reasoning}} \\
\hline
BLEU  & $^\dagger$\underline{0.1613} & 0.0100 & 0.0046 & \textbf{0.1620}  & 0.1374\\
ROUGE & $^\dagger$0.3706 & 0.1431 & 0.0924 & \underline{0.3720}  & \textbf{0.3873}\\
BERT  & $^\dagger$\textbf{0.9056} & 0.8682 & 0.8619 & \underline{0.9053} & \textbf{0.9056}\\
\hline
\multicolumn{6}{|l|}{\textit{Prevention Reasoning}} \\
\hline
BLEU  & 0.0030 & $^\dagger$\underline{0.0052} & 0.0030 & 0.0030 & \textbf{0.0173}\\
ROUGE & $^\dagger$\underline{0.3712} & 0.1409 & 0.3699 & \underline{0.3712} & \textbf{0.3800}\\
BERT  & $^\dagger$\underline{0.8920} & 0.8501 & $^\dagger$\underline{0.8920} & \underline{0.8920} & \textbf{0.8929}\\
\hline
\multicolumn{6}{|l|}{\textit{Crash Localization}} \\
\hline
mIoU  & 0.1636 & $^\dagger$\underline{0.1752} & 0.0474 & 0.0485 & \textbf{0.1897}\\
AP@30 & 0.1885 & $^\dagger$\underline{0.2496} & 0.0706 & 0.0671 & \textbf{0.2883}\\
AP@50 & 0.0602 & $^\dagger$\underline{0.0990} & 0.0267 & 0.0284 & \textbf{0.1368}\\
AP@70 & 0.0129 & $^\dagger$\textbf{0.0310} & 0.0060 & 0.0095 & \underline{0.0284}\\
\hline
\multicolumn{6}{|l|}{\textit{Pre-crash Localization}} \\
\hline
mIoU
& $^\dagger$\underline{0.2418} & 0.2043 & 0.0338 & 0.0342 & \textbf{0.2638}\\
AP@30
& $^\dagger$\underline{0.2995} & 0.2410 & 0.0491 & 0.0491 & \textbf{0.3348}\\
AP@50
& $^\dagger$\underline{0.2435} & 0.2117 & 0.0293 & 0.0301 & \textbf{0.2556}\\
AP@70
& $^\dagger$\textbf{0.1850} & 0.1670 & 0.0189 & 0.0207 & \underline{0.1764}\\
\hline
\end{tabularx}
\label{tab:sota_mllm}
\end{table}

Among the three 7B MLLMs, VideoLLaMA3 achieves the strongest and most balanced performance across both linguistic-centric and perception-centric tasks. It consistently outperforms LLaMA-ViD (7B) by margins of up to 0.278, with the sole exception of the BERT score for crash description, which decreases marginally by 0.018. VideoLLaMA3 also outperforms TimeChat on all tasks except for crash localization, with particularly pronounced advantages on linguistic-centric tasks. Although TimeChat is designed for time-sensitive long-video understanding and outperforms VideoLLaMA3 on crash localization, VideoLLaMA3 demonstrates superior capability in pre-crash localization, a more challenging temporal grounding problem that requires precise identification of early abnormal cues.

Beyond the 7B models, VideoLLaMA3 significantly outperforms the mid-size model LLaMA-ViD (13B) on crash recognition and temporal grounding tasks, with performance gains of up to 0.25, highlighting its stronger ability to model and detect fine-grained temporal dynamics in visual data. While Qwen3-VL (32B) achieves higher performance than VideoLLaMA3 across all evaluated tasks, the average improvement is limited to 0.019, which does not justify the substantially increased model size. This comparison further supports VideoLLaMA3 as a favorable baseline that offers a strong balance between performance and computational efficiency.

\subsection{Multitask Training Strategies}

\Cref{subsec: Training Multitask Models} proposes training two models with distinct task emphases. One model jointly optimizes the perception-centric tasks, namely crash localization and pre-crash localization. The other model jointly optimizes the linguistic-centric tasks, including recognition, description, causal reasoning, and prevention reasoning, while incorporating the perception-centric tasks as auxiliary objectives to facilitate cross-task knowledge transfer.

To assess the validity of this design, we conduct a comparative study across the following model configurations:

\begin{itemize}
    \item[\textbullet] \textbf{Independent monotask modelss}: Six monotask models, each trained on a single task, providing a baseline without shared representations.
    \item[\textbullet] \textbf{Homogeneous multitask models}: Two homogeneous multitask models: one trained on the four linguistic-centric tasks and the other on the two perception-centric tasks, designed to evaluate intra-group knowledge sharing.
    \item[\textbullet] \textbf{Heterogeneous multitask model}: One multitask model jointly trained on all six tasks, used to analyze cross-group transfer and potential negative interference.
\end{itemize}

\Cref{tab:decoupling_grouping_all} shows that multitask models consistently outperform their monotask counterparts on linguistic-centric tasks, achieving either the best or second-best performance across all metrics, with improvements of up to 0.0906. Furthermore, compared to the homogeneous multitask model, the heterogeneous multitask model yields gains of up to 0.0439. The only exceptions are the ROUGE and BERT scores for the prevention reasoning task, which exhibit negligible decreases of less than 0.0019. These results validate the heterogeneous model design, in which perception-centric tasks are incorporated as auxiliary objectives to enhance the learning of linguistic-centric tasks.

\begin{table}[ht]
\centering
\caption{
Evaluation of CrashChat's task decoupling and grouping strategy for both linguistic-centric and perception-centric tasks.
\textbf{Bold} values indicate the best performance, while \underline{underlined} values indicate the second-best performance.
}
\label{tab:decoupling_grouping_all}

\begin{tabularx}{\textwidth}{|
    >{\raggedright\arraybackslash}p{1.6cm}|
    >{\raggedleft\arraybackslash}p{1.45cm}
    >{\raggedleft\arraybackslash}p{1.45cm}
    >{\raggedleft\arraybackslash}p{1.45cm}|
    >{\raggedleft\arraybackslash}X
    >{\raggedleft\arraybackslash}X
    >{\raggedleft\arraybackslash}X|
}
\hline
& Ind.
& Homo.
& Hete.
& Homo.-Ind.
& Hete.-Ind.
& Hete.-Homo. \\
\hline

\multicolumn{7}{|l|}{\textit{Crash Recognition}} \\
\hline
Rec
& 0.8839
& \underline{0.9306}
& \textbf{0.9745}
& $\uparrow 0.047$
& $\uparrow 0.091$
& $\uparrow 0.044$ \\
Pre
& 0.9395
& \underline{0.9621}
& \textbf{0.9850}
& $\uparrow 0.023$
& $\uparrow 0.046$
& $\uparrow 0.023$ \\
F1
& 0.9024
& \underline{0.9433}
& \textbf{0.9793}
& $\uparrow 0.041$
& $\uparrow 0.077$
& $\uparrow 0.036$ \\
\hline

\multicolumn{7}{|l|}{\textit{Crash Description}} \\
\hline
BLEU
& 0.4320
& \underline{0.4421}
& \textbf{0.4608}
& $\uparrow 0.010$
& $\uparrow 0.029$
& $\uparrow 0.019$ \\
ROUGE
& 0.8175
& \underline{0.8366}
& \textbf{0.8535}
& $\uparrow 0.019$
& $\uparrow 0.036$
& $\uparrow 0.017$ \\
BERT
& 0.9736
& \underline{0.9760}
& \textbf{0.9783}
& $\uparrow 0.002$
& $\uparrow 0.005$
& $\uparrow 0.002$ \\
\hline

\multicolumn{7}{|l|}{\textit{Causal Reasoning}} \\
\hline
BLEU
& 0.2568
& \underline{0.2729}
& \textbf{0.3158}
& $\uparrow 0.016$
& $\uparrow 0.059$
& $\uparrow 0.043$ \\
ROUGE
& 0.5285
& \underline{0.5304}
& \textbf{0.5683}
& $\uparrow 0.002$
& $\uparrow 0.040$
& $\uparrow 0.038$ \\
BERT
& 0.9244
& \underline{0.9253}
& \textbf{0.9326}
& $\uparrow 0.001$
& $\uparrow 0.008$
& $\uparrow 0.007$ \\
\hline

\multicolumn{7}{|l|}{\textit{Prevention Reasoning}} \\
\hline
BLEU
& 0.1882
& \underline{0.1970}
& \textbf{0.2380}
& $\uparrow 0.009$
& $\uparrow 0.050$
& $\uparrow 0.041$ \\
ROUGE
& 0.5386
& \textbf{0.5568}
& \underline{0.5559}
& $\uparrow 0.018$
& $\uparrow 0.017$
& $\downarrow 0.001$ \\
BERT
& 0.9227
& \textbf{0.9268}
& \underline{0.9249}
& $\uparrow 0.004$
& $\uparrow 0.002$
& $\downarrow 0.002$ \\
\hline

\multicolumn{7}{|l|}{\textit{Crash Localization}} \\
\hline
mIoU
& \underline{0.5344}
& \textbf{0.5456}
& 0.4608
& $\uparrow 0.011$
& $\downarrow 0.074$
& $\downarrow 0.085$ \\
AP@30
& \underline{0.7883}
& \textbf{0.8003}
& 0.7031
& $\uparrow 0.012$
& $\downarrow 0.085$
& $\downarrow 0.097$ \\
AP@50
& \underline{0.5749}
& \textbf{0.5998}
& 0.4785
& $\uparrow 0.025$
& $\downarrow 0.096$
& $\downarrow 0.121$ \\
AP@70
& \underline{0.3201}
& \textbf{0.3348}
& 0.2229
& $\uparrow 0.015$
& $\downarrow 0.097$
& $\downarrow 0.112$ \\
\hline

\multicolumn{7}{|l|}{\textit{Pre-crash Localization}} \\
\hline
mIoU
& \underline{0.4351}
& \textbf{0.5018}
& 0.3148
& $\uparrow 0.067$
& $\downarrow 0.120$
& $\downarrow 0.187$ \\
AP@30
& \underline{0.5207}
& \textbf{0.5886}
& 0.3726
& $\uparrow 0.068$
& $\downarrow 0.148$
& $\downarrow 0.216$ \\
AP@50
& \underline{0.4518}
& \textbf{0.5293}
& 0.3305
& $\uparrow 0.078$
& $\downarrow 0.121$
& $\downarrow 0.199$ \\
AP@70
& \underline{0.3761}
& \textbf{0.4355}
& 0.2668
& $\uparrow 0.059$
& $\downarrow 0.109$
& $\downarrow 0.169$ \\
\hline
\end{tabularx}
\end{table}

On perception-centric tasks, \Cref{tab:decoupling_grouping_all} shows that the homogeneous multitask model achieves the best performance, outperforming the monotask counterparts (i.e., the second-best) by up to 0.0775. This result confirms the benefit of jointly learning two closely related tasks that both focus on identifying critical temporal turning points. In contrast, the heterogeneous multitask model exhibits the worst performance, with score reductions ranging from 0.0848 to 0.2160 comparing to the homogeneous multitask model. This degradation suggests substantial negative interference from linguistic-centric tasks, thereby confirming the necessity of task decoupling for effective perception-centric learning.

\subsection{Effectiveness in Crash Video Analysis}

\begin{figure}[ht]
    \centering
    \includegraphics[width=\linewidth]{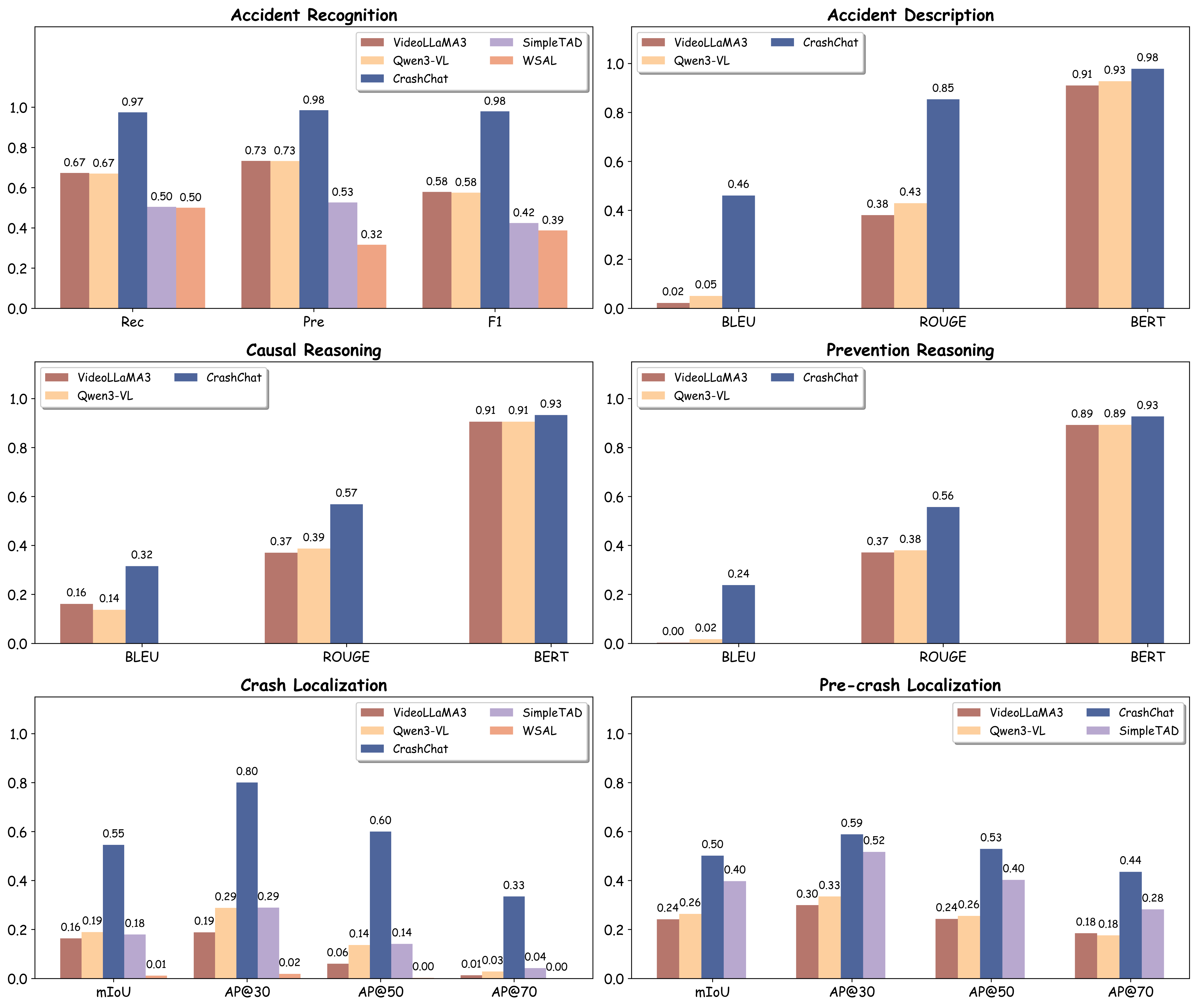}
    \caption{Comprehensive performance comparison of CrashChat, traditional vision-based models, and MLLMs across six core crash video analysis tasks}
    \label{fig3}
\end{figure}

To evaluate the effectiveness of domain adaptation, we compare CrashChat with two traditional vision-based baselines (SimpleTAD and WSAL) as well as two representative MLLMs (VideoLLaMA3 and Qwen3-VL) across six core crash video analysis tasks. The results are summarized in Fig.~\ref{fig3}, which 
indicates that instruction fine-tuning combined with proposed multitask learning can effectively inject traffic-crash domain knowledge into VideoLLaMA3, transforming a general-purpose MLLM into a specialized crash analysis model.

Across all tasks, CrashChat achieves the best overall performance. For crash recognition, CrashChat attains near-saturated accuracy, yielding up to 0.30 improvement on F1 score over MLLM models (VideoLLaMA3 and Qwen3-VL) and 0.47 over vision-based approaches (SimpleTAD and SWAL). In crash description and reasoning tasks, CrashChat shows  0.16$\sim$0.44 improvements in the BLEU score and 0.19$\sim$0.47 in ROUGE scores, reflecting enhanced textual accuracy and content coverage. The BERT scores increase by approximately 0.02$\sim$0.07, indicating that semantic alignment is preserved and modestly improved.

In perception-centric tasks, CrashChat demonstrates clear superiority over all prior models. For crash localization, CrashChat achieves an mIoU of 0.55, corresponding to up to 0.36 improvement over the MLLM models and up to 0.37 over the vision-based models. For the more challenging pre-crash localization task, CrashChat achieves 0.50 mIoU, continuing to dominate other models with margins ranging from 0.1$\sim$0.26. WSAL fails to generalize to pre-crash anticipation and does not yield meaningful localization results.

We conclude that CrashChat not only outperforms existing MLLMs but also surpasses traditional vision-based models, further validating the feasibility and effectiveness of adapting MLLMs to traffic crash video analysis.

\subsection{Qualitative Examples}

\begin{figure}
\includegraphics[width=\textwidth]{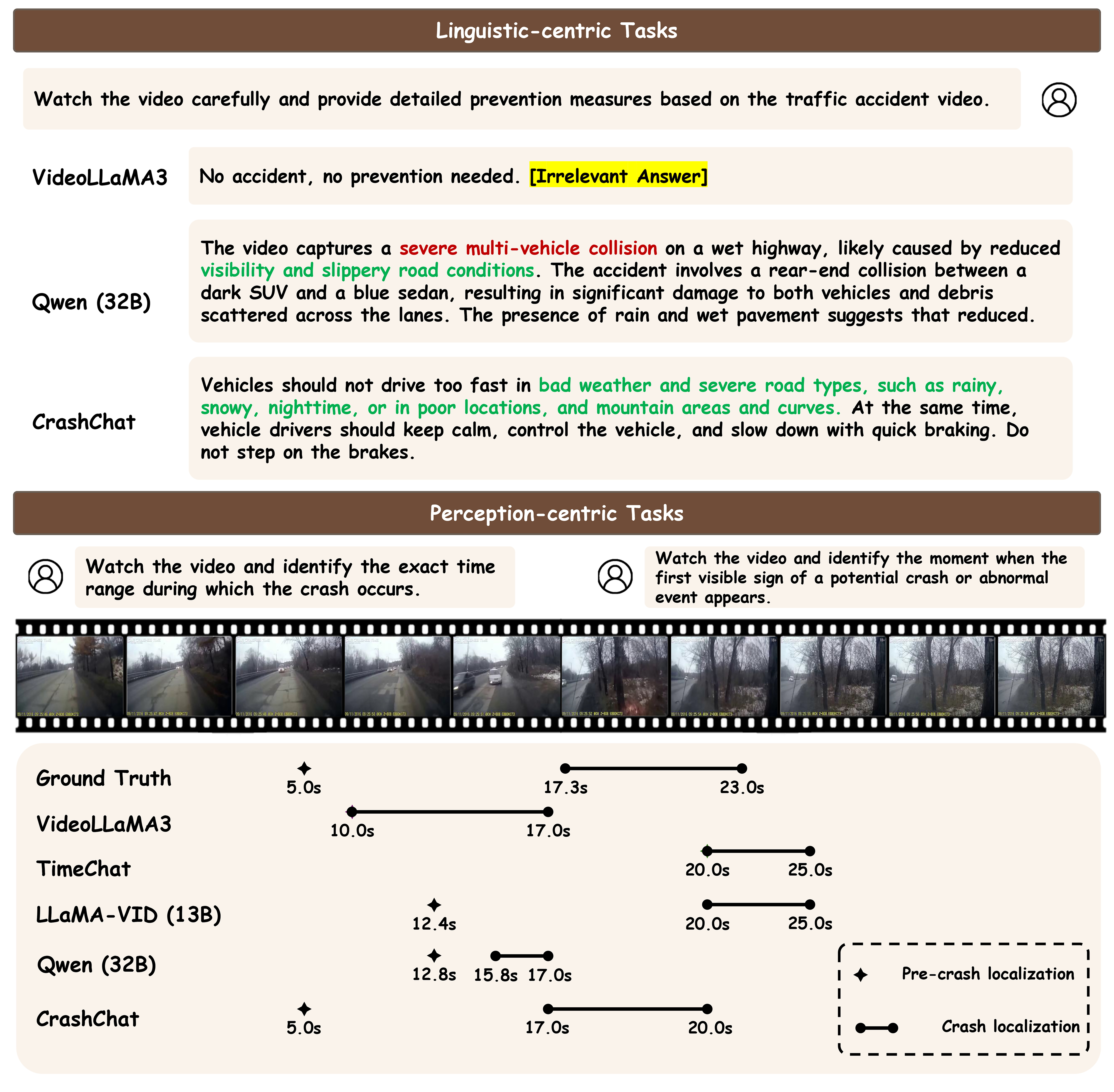}
\caption{
Qualitative comparison of linguistic-centric and perception-centric tasks.
\textcolor{CrashRed}{Red} highlights key semantic inconsistencies, and
\textcolor{CrashGreen}{Green} indicates strong alignment with the ground truth.}\label{fig4}
\end{figure}

In addition to the quantitative analysis, we further illustrate an example in Fig.~\ref{fig4} to illustrate CrashChat’s competitive performance in crash video analysis tasks. As shown in the figure, general MLLMs exhibit notable limitations: VideoLLaMA3 fails to recognize the presence of a crash and thus generates incorrect response; Qwen3-VL does not produce task-relevant answers and instead generates hallucinated content that deviates from the desired output. In contrast, CrashChat accurately understands the traffic scene and produces complete, instruction-aligned prevention reasoning results, demonstrating its strong capability in linguistic-centric tasks.

CrashChat also demonstrates strong performance on perception-centric tasks. It accurately identifies the start and end timestamps of crashes and localizes the earliest moment at which crash risk becomes visually detectable in the video. Compared to other general-purpose MLLMs, the instruction-tuned CrashChat is more effective at distinguishing fine-grained, time-sensitive moments associated with crash events.

\section{Conclusion}
\label{sec:Concusion}

This paper presents CrashChat, a MLLM designed for multitask traffic crash video analysis. Built on VideoLLaMA3, CrashChat combines instruction fine-tuning with task-aware multitask learning to effectively integrate linguistic reasoning and spatiotemporal perception. Comprehensive experiments show that CrashChat consistently outperforms existing MLLMs and traditional vision-based methods across six core crash analysis tasks, achieving SOTA performance. Further analysis demonstrates that task decoupling is critical for improving perception-centric tasks, while heterogeneous multitask learning benefits language-centric reasoning. Qualitative results confirm that CrashChat produces accurate, instruction-aligned responses with reliable temporal grounding, reducing hallucinations and missed detections. Overall, this work highlights the effectiveness of domain adaption and multitasking for applying MLLMs to traffic crash video analysis.

%
%
\bibliographystyle{splncs04}
\bibliography{references}

@article{accidentgpt_2023,
  author  = {Wang, L. and Ren, Y. and Jiang, H. and Cai, P. and Fu, D. and Wang, T. and Wang, Y.},
  title   = {AccidentGPT: Accident Analysis and Prevention from V2X Environmental Perception with Multi-modal Large Model},
  journal = {arXiv preprint arXiv:2312.13156},
  year    = {2023}
}

@inproceedings{106k_2025,
  author    = {Zhou, Y. and Bai, L. and Cai, S. and Deng, B. and Xu, X. and Shen, H.~T.},
  title     = {TAU-106K: A New Dataset for Comprehensive Understanding of Traffic Accident},
  booktitle = {Proceedings of the International Conference on Learning Representations (ICLR)},
  year      = {2025}
}

@article{safeplug_2025,
  author  = {Sheng, Z. and Huang, Z. and Qu, Y. and Chen, J. and Luo, Y. and Chen, Y.~J. and Chen, S.},
  title   = {SafePLUG: Empowering Multimodal LLMs with Pixel-Level Insight and Temporal Grounding for Traffic Accident Understanding},
  journal = {arXiv preprint arXiv:2508.06763},
  year    = {2025}
}

@inproceedings{echo_2025,
  author    = {Xing, Z. and Chen, H. and Xie, B. and Xu, J. and Guo, Z. and Xu, X. and Heng, P.~A.},
  title     = {EchoTraffic: Enhancing Traffic Anomaly Understanding with Audio-Visual Insights},
  booktitle = {Proceedings of the IEEE/CVF Conference on Computer Vision and Pattern Recognition (CVPR)},
  pages     = {19098--19108},
  year      = {2025}
}

@article{videollama3_2025,
  author  = {Zhang, B. and Li, K. and Cheng, Z. and Hu, Z. and Yuan, Y. and Chen, G. and Zhao, D.},
  title   = {VideoLLaMA 3: Frontier Multimodal Foundation Models for Image and Video Understanding},
  journal = {arXiv preprint arXiv:2501.13106},
  year    = {2025}
}

@inproceedings{yao2019unsupervised,
  author    = {Yao, Y. and Xu, M. and Wang, Y. and Crandall, D.~J. and Atkins, E.~M.},
  title     = {Unsupervised Traffic Accident Detection in First-Person Videos},
  booktitle = {IEEE/RSJ International Conference on Intelligent Robots and Systems (IROS)},
  pages     = {273--280},
  year      = {2019}
}

@inproceedings{fang2019dada,
  author    = {Fang, J. and Yan, D. and Qiao, J. and Xue, J. and Wang, H. and Li, S.},
  title     = {DADA-2000: Can Driving Accident be Predicted by Driver Attention?},
  booktitle = {IEEE Intelligent Transportation Systems Conference (ITSC)},
  pages     = {4303--4309},
  year      = {2019}
}

@article{lv2021wsal,
  author  = {Lv, H. and Zhou, C. and Cui, Z. and Xu, C. and Li, Y. and Yang, J.},
  title   = {Localizing Anomalies from Weakly-Labeled Videos},
  journal = {IEEE Transactions on Image Processing},
  volume  = {30},
  pages   = {4505--4515},
  year    = {2021}
}

@article{yao2022dota,
  author  = {Yao, Y. and Wang, X. and Xu, M. and Pu, Z. and Wang, Y. and Atkins, E. and Crandall, D.~J.},
  title   = {DoTA: Unsupervised Detection of Traffic Anomaly in Driving Videos},
  journal = {IEEE Transactions on Pattern Analysis and Machine Intelligence},
  volume  = {45},
  number  = {1},
  pages   = {444--459},
  year    = {2022}
}

@article{karim2023amnet,
  author  = {Karim, M.~M. and Yin, Z. and Qin, R.},
  title   = {An Attention-Guided Multistream Feature Fusion Network for Early Localization of Risky Traffic Agents in Driving Videos},
  journal = {IEEE Transactions on Intelligent Vehicles},
  volume  = {9},
  number  = {1},
  pages   = {1792--1803},
  year    = {2023}
}

@inproceedings{fang2024abductive,
  author    = {Fang, J. and Li, L.~L. and Zhou, J. and Xiao, J. and Yu, H. and Lv, C. and Chua, T.~S.},
  title     = {Abductive Ego-View Accident Video Understanding for Safe Driving Perception},
  booktitle = {IEEE/CVF Conference on Computer Vision and Pattern Recognition (CVPR)},
  pages     = {22030--22040},
  year      = {2024}
}

@inproceedings{moura2025nexar,
  author    = {Moura, D. and Zhu, S. and Zvitia, O.},
  title     = {Nexar Dashcam Collision Prediction Dataset and Challenge},
  booktitle = {IEEE/CVF Conference on Computer Vision and Pattern Recognition (CVPR)},
  pages     = {2583--2591},
  year      = {2025}
}

@article{d2city,
  author  = {Che, Z. and Li, G. and Li, T. and Jiang, B. and Shi, X. and Zhang, X. and Ye, J.},
  title   = {D$^{2}$-City: A Large-Scale Dashcam Video Dataset of Diverse Traffic Scenarios},
  journal = {arXiv preprint arXiv:1904.01975},
  year    = {2019}
}

@inproceedings{lora,
  author    = {Hu, E.~J. and Shen, Y. and Wallis, P. and Allen-Zhu, Z. and Li, Y. and Wang, S. and Chen, W.},
  title     = {LoRA: Low-Rank Adaptation of Large Language Models},
  booktitle = {International Conference on Learning Representations (ICLR)},
  year      = {2022}
}

@inproceedings{ding_2023_etrnlp,
  author    = {Ding, Jianyu and Li, Xudong and Wang, Wei and Tang, Hao and Wang, Zicheng and Deng, Jia},
  title     = {Mitigating Task Interference in Multi-Task Learning via Explicit Task Routing with Non-Learnable Primitives},
  booktitle = {Proceedings of the IEEE/CVF Conference on Computer Vision and Pattern Recognition (CVPR)},
  pages     = {21455--21464},
  year      = {2023}
}

@inproceedings{standley2020which,
  author    = {Standley, Trevor and Zamir, Amir R. and Chen, Dawn and Guibas, Leonidas and Malik, Jitendra and Savarese, Silvio},
  title     = {Which Tasks Should Be Learned Together in Multi-Task Learning?},
  booktitle = {Proceedings of the International Conference on Machine Learning (ICML)},
  pages     = {9120--9132},
  year      = {2020},
  publisher = {PMLR}
}

@article{fang2022cognitive,
  author  = {Fang, J. and Li, L.-L. and Yang, K. and Zheng, Z. and Xue, J. and Chua, T.-S.},
  title   = {Cognitive Accident Prediction in Driving Scenes: A Multimodality Benchmark},
  journal = {CoRR},
  volume  = {abs/2212.09381},
  year    = {2022}
}

@inproceedings{orlova2025simplifying,
  author    = {Orlova, Svetlana and Kerssies, Tommie and Englert, Brun{\'o} B and Dubbelman, Gijs},
  title     = {Simplifying Traffic Anomaly Detection with Video Foundation Models},
  booktitle = {Proceedings of the IEEE/CVF International Conference on Computer Vision (ICCV)},
  pages     = {852--862},
  year      = {2025},
  publisher = {IEEE}
}

@article{liang2024textdriven,
  author  = {Liang, R. and Li, Y. and Zhou, J. and Li, X.},
  title   = {Text-driven Traffic Anomaly Detection with Temporal High-frequency Modeling in Driving Videos},
  journal = {IEEE Transactions on Circuits and Systems for Video Technology},
  volume  = {34},
  number  = {9},
  pages   = {8684--8697},
  year    = {2024}
}

@article{sun2025earccpmnet,
  author  = {Sun, W. and Abdullah, L. N. and Khalid, F. B. and Sulaiman, P. Suhaiza Binti},
  title   = {EAR-CCPM-Net: A Cross-Modal Collaborative Perception Network for Early Accident Risk Prediction},
  journal = {Applied Sciences},
  volume  = {15},
  number  = {17},
  pages   = {9299},
  year    = {2025}
}

@article{jaradat2025near_miss,
  author  = {Jaradat, S. and Elhenawy, M. and Ashqar, H.~I. and Paz, A. and Nayak, R.},
  title   = {Leveraging Deep Learning and Multimodal Large Language Models for Near-Miss Detection Using Crowdsourced Videos},
  journal = {IEEE Open Journal of the Computer Society},
  year    = {2025}
}

@article{yao2025ifinder,
  author  = {Yao, M. and Zhuang, B. and Garg, S. and Roy-Chowdhury, A. and Shelton, C. and Chandraker, M. and Aich, A.},
  title   = {iFinder: Structured Zero-Shot Vision-Based LLM Grounding for Dash-Cam Video Reasoning},
  journal = {arXiv preprint arXiv:2509.19552},
  year    = {2025}
}

@inproceedings{shi2025scvlm,
  author    = {Shi, L. and Jiang, B. and Zeng, T. and Guo, F.},
  title     = {ScVLM: Enhancing Vision-Language Model for Safety-Critical Event Understanding},
  booktitle = {Proceedings of the IEEE/CVF Winter Conference on Applications of Computer Vision (WACV)},
  pages     = {1061--1071},
  year      = {2025}
}

@inproceedings{li_ictd_2023_gaze,
  author    = {Li, Yu and Karim, Muhammad Monjurul and Qin, Ruwen},
  title     = {A Gaze Data-Based Comparative Study to Build a Trustworthy Human--AI Collaboration in Crash Anticipation},
  booktitle = {Proceedings of the International Conference on Transportation and Development (ICTD)},
  year      = {2023}
}

\end{document}